\PassOptionsToPackage{table}{xcolor}
\documentclass[sigconf,nonacm]{acmart}

\usepackage[table]{xcolor}
\usepackage{graphicx}
\usepackage{amsmath}
\usepackage{multirow}
\usepackage{booktabs}
\usepackage{url}
\usepackage[unicode]{hyperref}

\usepackage{tikz}
\usetikzlibrary{positioning,arrows.meta}

\setcopyright{none}
\acmDOI{}
\acmISBN{}
\acmPrice{}
\acmConference[]{}{}{}

\begin{document}


\title{A Synthetic Multi-Modal Testbed for Security Stress Testing
of Agentic Counter-UAS Systems}


\author{Ivan A. Fernandez}
\affiliation{%
  \institution{Mississippi State University}
  \country{Mississippi State, MS, USA}
}
\email{iaf28@msstate.edu}

\author{Jingdao Chen}
\affiliation{%
  \institution{Mississippi State University}
  \country{Mississippi State, MS, USA}
}
\email{chenjingdao@cse.msstate.edu}


\begin{abstract}
Agentic cyber-physical systems rely on perception and state estimation
to support autonomous decisions. Degraded, inconsistent, or
manipulated sensor observations can therefore propagate through an
autonomous pipeline and create security-relevant failures. This work
presents a synthetic Counter-Unmanned Aircraft System (C-UAS) testbed
developed in NVIDIA Isaac Sim for repeatable stress testing of
perception and tracking algorithms supporting agentic decision-making.
A simplified quadrotor follows scripted trajectories while co-located
virtual sensors generate synchronized RGB, RTX LiDAR, and illustrative
IR-like observations. The testbed varies target range, motion, clutter,
occlusion, and sensor availability, with future stress cases including
sensor dropout, temporal misalignment, cross-modal disagreement, and
false observations. As an initial exemplar, a nonlinear coordinated-turn
Extended Kalman Filter (EKF) is evaluated over a 20-s maneuvering
trajectory using LiDAR target-centroid measurements. Position error and
uncertainty are compared as LiDAR observations degrade and are lost.
The framework provides a reproducible environment for studying how
sensing degradation affects downstream autonomous estimation and
resilience in agentic C-UAS systems.
\end{abstract}

\maketitle



\section{Introduction \& Motivation}

Agentic systems increasingly combine perception, state estimation,
reasoning, and autonomous action to interact with dynamic
environments \cite{pati2025agentic,sapkota2025agentic}. In
cyber-physical applications, degraded or manipulated sensor
observations can therefore influence downstream autonomous decisions.
This concern is particularly relevant to Counter-Unmanned Aircraft
System (C-UAS) architectures, where detection, tracking,
classification, and sensor fusion may support subsequent decision and
response functions. Recent U.S. defense requirements similarly
emphasize robust C-UAS sensing, tracking, multi-target handling, and
operation under challenging conditions
\cite{usarmy2026mobilecsuas,usaf2026csuas}. The operational relevance
of drone and counter-drone technologies has also increased
substantially in recent conflicts \cite{mittal2025drone}.

Real-world datasets remain essential for evaluating C-UAS sensing and
tracking; however, they provide limited control over the occurrence,
severity, and repeatability of individual stress conditions.
Standardized C-UAS evaluation efforts have likewise emphasized
scenario-based testing, repeatability, and quantitative performance
metrics \cite{decubber2025standardized}. Multi-modal datasets such as
MMAUD provide synchronized camera, LiDAR, radar, and other sensing
data for anti-UAV research \cite{yuan2024mmaud}, while synthetic
datasets such as SynDroneVision demonstrate the value of simulation
for increasing scenario diversity and supplementing real-world data
\cite{lenhard2025syndronevision}.

This work develops a lightweight synthetic C-UAS testbed for security
stress testing of perception and tracking components that may feed
agentic decision pipelines. The objective is not to reproduce
high-fidelity aerodynamics or fully calibrated sensor physics.
Instead, the testbed emphasizes repeatable target motion, known ground
truth, configurable scene conditions, and controlled sensing
degradation so that downstream algorithm behavior can be observed
under systematically varied conditions.


\section{Synthetic Multi-Modal Testbed}

\begin{figure}[t]
\centering

\begin{minipage}[c]{0.30\columnwidth}
\centering

\begin{tikzpicture}[
    node distance=2.5mm,
    box/.style={
        draw,
        rounded corners=1.5pt,
        text width=1.8cm,
        minimum height=7mm,
        align=center,
        font=\scriptsize,
        inner sep=2.5pt
    },
    future/.style={
        draw,
        dashed,
        rounded corners=1.5pt,
        text width=1.8cm,
        minimum height=7mm,
        align=center,
        font=\scriptsize,
        inner sep=2.5pt
    },
    arrow/.style={
        -{Latex},
        line width=0.6pt
    }
]

\node[box] (stress) {
    \textbf{Scenario \& Stressors}\\
    Motion, occlusion,\\
    dropout, misalignment
};

\node[box, below=of stress] (sim) {
    \textbf{Isaac Sim}\\
    UAV + environment\\
    Ground truth
};

\node[box, below=of sim] (sensors) {
    \textbf{Synthetic Sensing}\\
    RGB\\
    RTX LiDAR\\
    IR-like
};

\node[box, below=of sensors] (tracking) {
    \textbf{Perception / Tracking}\\
    Detection\\
    Fusion\\
    Nonlinear EKF
};

\node[future, below=of tracking] (agent) {
    \textbf{Agentic Monitoring}\\
    Resilience\\
    Fallback / oversight
};

\draw[arrow] (stress) -- (sim);
\draw[arrow] (sim) -- (sensors);
\draw[arrow] (sensors) -- (tracking);
\draw[arrow, dashed] (tracking) -- (agent);

\end{tikzpicture}

\end{minipage}
\hfill
\begin{minipage}[c]{0.66\columnwidth}
\centering

{\small\textbf{Algorithm Analysis}}\\[-1mm]

\vspace{1mm}

\includegraphics[
    width=\linewidth
]{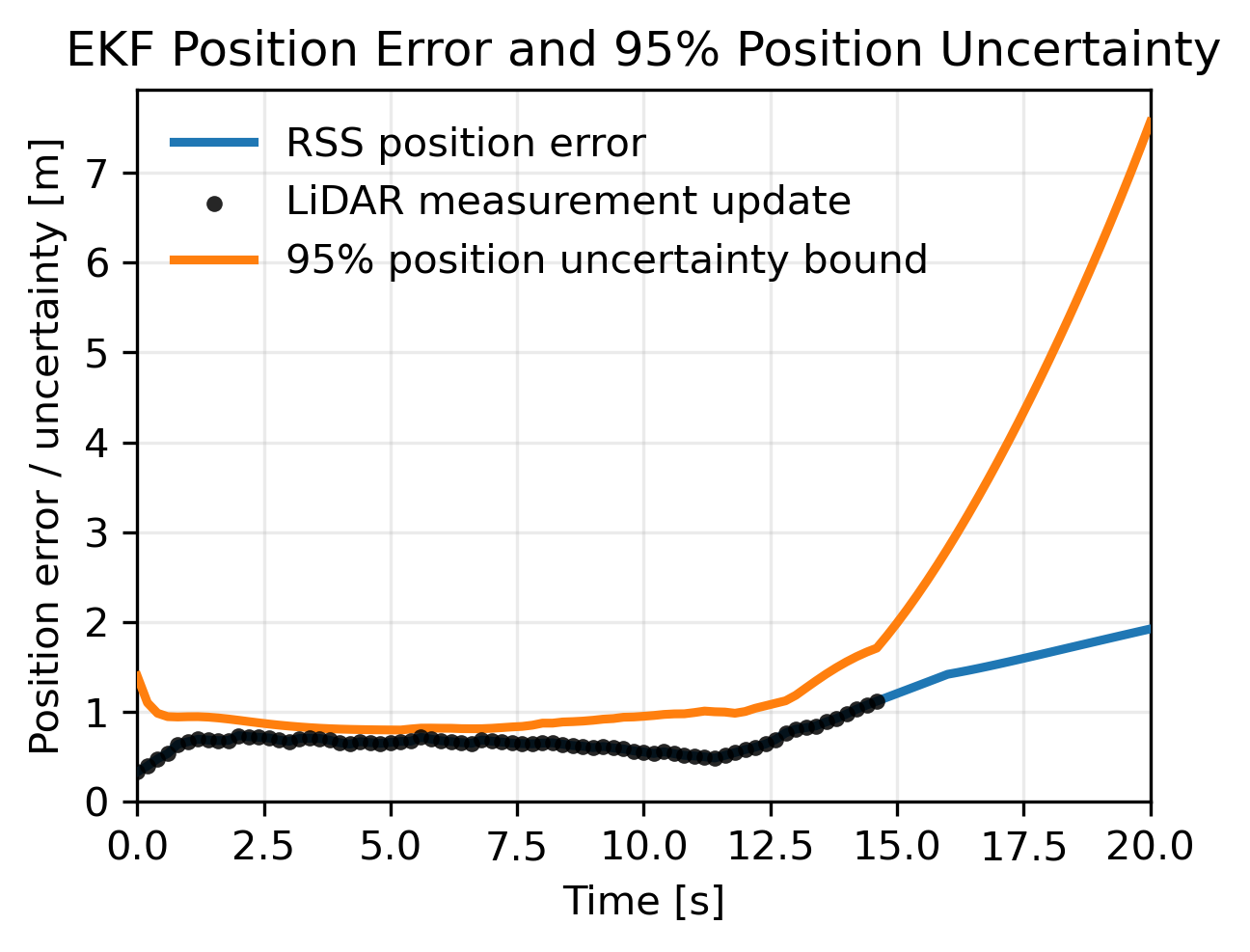}

\vspace{-2mm}

{\scriptsize Nonlinear EKF position error and uncertainty under
degraded LiDAR sensing.}

\end{minipage}

\vspace{-2mm}

\caption{Synthetic C-UAS security stress-testing framework and
tracking exemplar. Controlled sensing conditions are generated in
NVIDIA Isaac Sim for downstream perception and tracking evaluation.
The nonlinear EKF result illustrates position error and uncertainty
under degraded LiDAR sensing; agentic monitoring represents future
work.}

\label{fig:framework_analysis}

\vspace{-4mm}
\end{figure}

Figure~\ref{fig:framework_analysis} summarizes the synthetic testbed. Scenario
parameters specify target trajectory, range, sensor configuration, and
environmental conditions. A simplified quadrotor-shaped Universal
Scene Description (USD) target is propagated using prescribed
kinematic trajectories, while Isaac Sim generates synchronized
multi-modal observations together with known target state and sensor
geometry.

\begin{figure*}[t]
\centering
\setlength{\tabcolsep}{3pt}

\begin{tabular}{cccc}

\multicolumn{2}{c}{\textbf{Near / Clear}} &
\multicolumn{2}{c}{\textbf{Clutter / Occlusion}}
\\[-1mm]

\includegraphics[
    width=0.235\textwidth,
    height=0.13\textheight,
    keepaspectratio
]{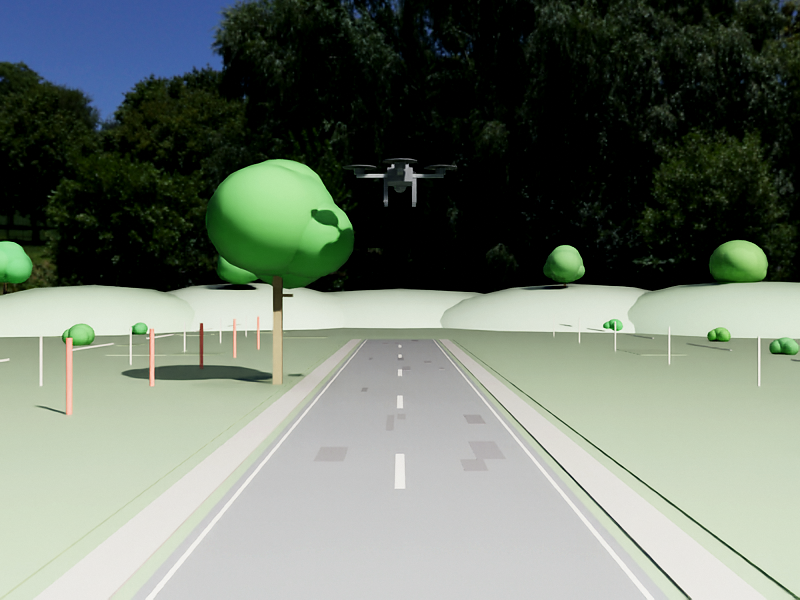}
&
\includegraphics[
    width=0.235\textwidth,
    height=0.13\textheight,
    keepaspectratio
]{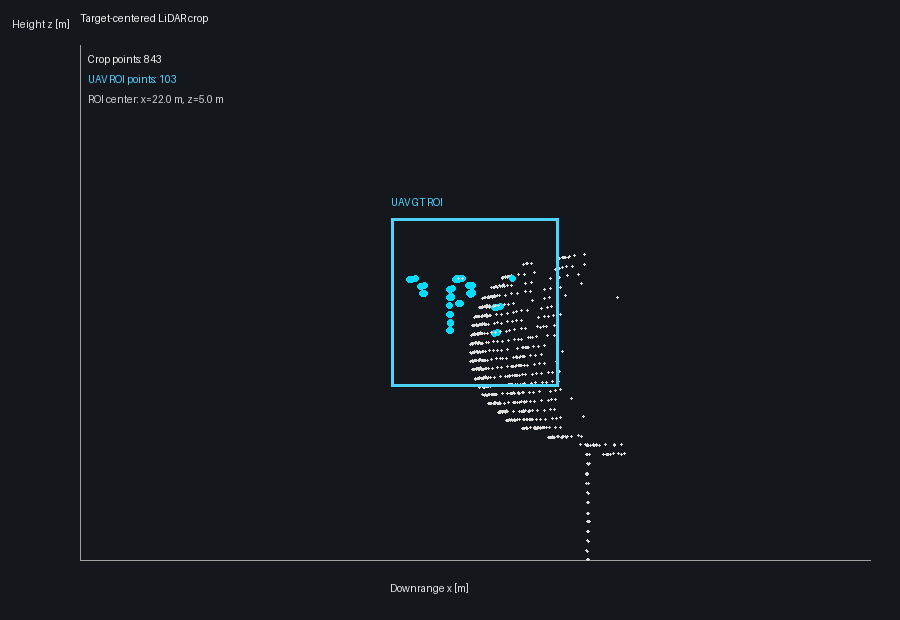}
&
\includegraphics[
    width=0.235\textwidth,
    height=0.13\textheight,
    keepaspectratio
]{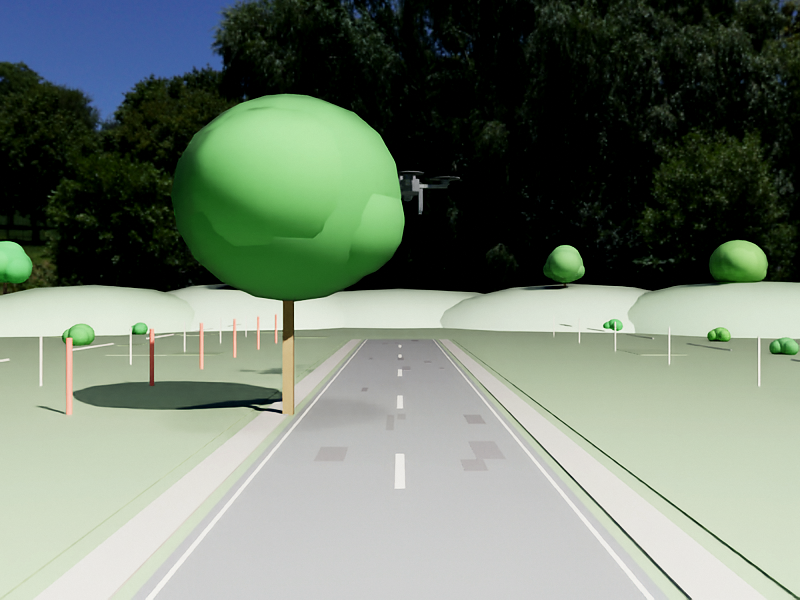}
&
\includegraphics[
    width=0.235\textwidth,
    height=0.13\textheight,
    keepaspectratio
]{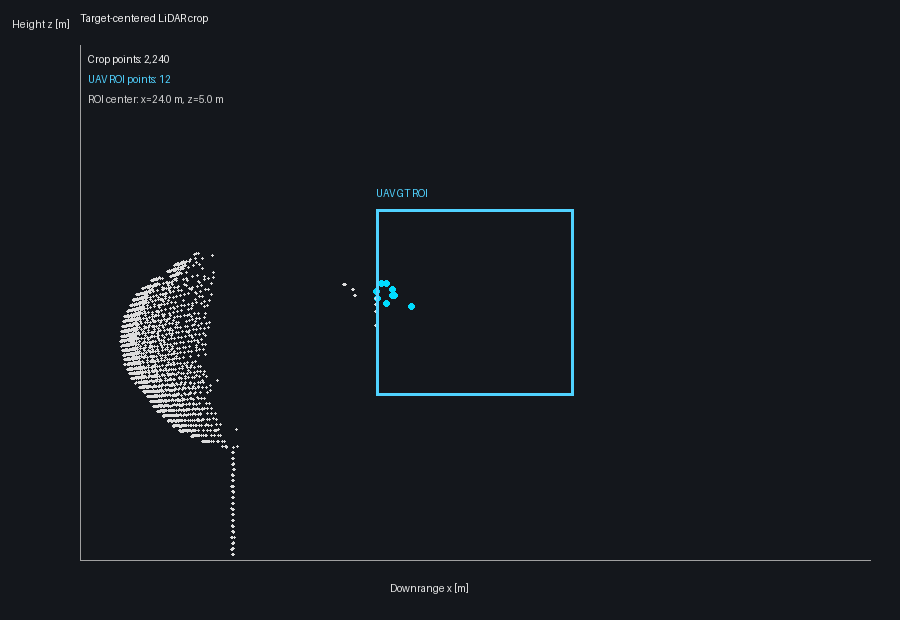}
\\[-1mm]

{\scriptsize RGB} &
{\scriptsize RTX LiDAR} &
{\scriptsize RGB} &
{\scriptsize RTX LiDAR}

\end{tabular}

\vspace{-2mm}

\caption{Representative synthetic C-UAS sensing scenarios generated
using NVIDIA Isaac Sim. Paired RGB and RTX LiDAR observations compare
a near-range clear condition with a cluttered and partially occluded
target condition. The testbed additionally supports illustrative
IR-like observations for multi-modal experiments.}

\label{fig:exemplars}

\vspace{-4mm}
\end{figure*}

The target follows repeatable trajectories including straight flight,
hover, climb or descent, acceleration, and turning motion. Because
target state is known directly from the simulation, generated
observations can be paired with ground truth for quantitative
evaluation.

The RGB camera provides rendered imagery, while RTX LiDAR produces
three-dimensional returns from target and scene geometry. An
illustrative IR-like channel provides a third modality for preliminary
multi-modal experiments; it is intended as a thermal-like synthetic
representation rather than a physics-based long-wave infrared model.
Target range, viewing geometry, clutter, occlusion, and sensor
availability can therefore be varied while preserving a common
underlying trajectory.

From a security perspective, these controls provide a basis for
repeatable observation-level perturbations that may affect downstream
autonomous behavior. Candidate stress cases include sensor dropout,
reduced update rate, temporal misalignment, cross-modal disagreement,
false or ghost observations, and ambiguous multi-target encounters.
Such conditions can support studies of when uncertainty or
inconsistency should trigger monitoring, fallback behavior, or human
oversight.


\section{Synthetic Data Exemplars}

Three representative sensing configurations demonstrate the current
testbed: a near-range target in a relatively clear scene, an
increased-range target, and a target observed in clutter with partial
occlusion. Figure~\ref{fig:exemplars} shows paired RGB and RTX LiDAR
observations generated from the same underlying scene state.

These scenarios demonstrate controlled multi-modal data generation
rather than sensor-model fidelity. Future experiments will introduce
security-relevant sensing perturbations while retaining the same
ground-truth trajectory and evaluation structure.


\section{Nonlinear EKF Tracking Exemplar}

LiDAR-based UAV detection and tracking has previously been explored
using spatiotemporal point-cloud processing and filtering approaches
\cite{liang2024cluster,khosravi2026lidar}. Here, the EKF is used only
as an exemplar for demonstrating how the proposed testbed can expose a
tracker to controlled measurement degradation.

A nonlinear coordinated-turn EKF is evaluated over a 20-s maneuvering
trajectory using LiDAR target-centroid measurements. The filter state
is

\begin{equation}
\mathbf{x}_k =
[p_x,p_y,p_z,v,\psi,v_z,\omega]^\top ,
\end{equation}

where $p_x$, $p_y$, and $p_z$ denote target position, $v$ is
horizontal speed, $\psi$ is heading, $v_z$ is vertical velocity, and
$\omega$ is turn rate. The LiDAR measurement is represented as

\begin{equation}
\mathbf{z}_k = \mathbf{p}_k + \mathbf{e}_k ,
\end{equation}

where $\mathbf{p}_k=[p_x,p_y,p_z]^\top$ is the target position and
$\mathbf{e}_k$ is the LiDAR centroid measurement error. The
visible-surface centroid is therefore treated as an imperfect target
position measurement.

Figure~\ref{fig:framework_analysis} demonstrates quantitative tracking evaluation.
Uncertainty remains bounded during regular LiDAR updates and increases
as target returns become sparse and measurements are lost, illustrating
how sensing degradation propagates into estimator state and
uncertainty.


\vspace{-2mm}
\section{Conclusion \& Future Work}

This work presents an NVIDIA Isaac Sim testbed for controlled
generation of synthetic C-UAS observations and repeatable security
stress testing of perception and tracking algorithms. Scripted
quadrotor motion and synchronized RGB, RTX LiDAR, and illustrative
IR-like sensing support multi-modal evaluation. A nonlinear
coordinated-turn EKF demonstrates quantitative assessment of tracking
error and uncertainty under degraded sensing.

Future work will introduce sensor dropout, temporal misalignment,
cross-modal disagreement, false observations, occlusion,
environmental degradation, and multi-target interactions. These
scenarios will support studies of how perception and tracking failures
affect downstream agentic behavior and inform monitoring, resilience,
fallback, and human oversight.


\bibliographystyle{unsrt}
\bibliography{refs}

\end{document}